\pdfoutput=1
\documentclass[11pt]{article}

\usepackage{acl}
\usepackage{times}
\usepackage{latexsym}

\usepackage[T1]{fontenc}

\usepackage[utf8]{inputenc}
\usepackage{microtype}
\usepackage{listings}
\usepackage{fancyvrb}
\usepackage{graphicx}
\usepackage{multirow}
\usepackage{subcaption}
\usepackage{graphicx}
\usepackage{babel,blindtext}
\usepackage{enumitem}
\usepackage{colortbl}
\usepackage{caption}
\usepackage{hhline}

\title{ChatGPT for Suicide Risk Assessment on Social Media: Quantitative Evaluation of Model Performance, Potentials and Limitations}

\author{Hamideh Ghanadian{$^{1}$}, Isar Nejadgholi{$^{2}$},  Hussein Al Osman{$^3$} \\
  {$^{1,2,3}$}University of Ottawa, Ottawa, Canada\\
  {$^2$}National Research Council Canada, Ottawa, Canada \\
  \footnotesize\texttt{{$^{1,3}$}\{Hghan053, Hussein.alosman\}@uottawa.ca}\\   \footnotesize\texttt{{$^{2}$}isar.nejadgholi@nrc-cnrc.gc.ca}}

\begin{document}
\maketitle
\begin{abstract}

This paper presents a novel framework for quantitatively evaluating the interactive ChatGPT model in the context of  suicidality assessment from social media posts, utilizing the University of Maryland Reddit suicidality dataset. We conduct a technical evaluation of ChatGPT's performance on this task using Zero-Shot and Few-Shot experiments and compare its results with those of two fine-tuned transformer-based models. Additionally, we investigate the impact of different temperature parameters on ChatGPT's response generation and discuss the optimal temperature based on the inconclusiveness rate of ChatGPT. Our results indicate that while ChatGPT attains considerable accuracy in this task, transformer-based models fine-tuned on human-annotated datasets exhibit superior performance. Moreover, our  analysis sheds light on how adjusting the ChatGPT's hyperparameters can improve its ability to assist mental health professionals in this critical task. 
\end{abstract}

\section{Introduction}\label{sec:intro}
According to the World Health Organization (WHO)\footnote{\href{https://www.who.int/news-room/fact-sheets/detail/suicide}{The World Health Organization}}, more than 700,000 people die due to suicide every year. For every suicide, there are many more people who attempt suicide. Furthermore, suicide is the fourth leading cause of death among 15-29 year-olds. According to The US Centers for Disease Control and Prevention (CDC) \footnote{\href{https://www.cdc.gov/nchs/data/vsrr/vsrr024.pdf}{The US Centers for Disease Control and Prevention}}, the rate of suicides per 100,000 increased from 13.5 in 2020 to 14.0 in 2021. 

Social media platforms are becoming a common way for people to express their feelings, suffering, and suicidal tendencies.  One of the most effective methods recommended by the WHO for preventing suicide is to obtain information from social media and report suicidal ideation to healthcare providers to enable early identification, assessment, and follow-up with affected individuals.  Hence, social media provides a significant means for obtaining information to identify individuals who are at risk of committing suicide, allowing for timely detection and intervention \citep{abdulsalam2022suicidal}

In recent years, there has been a growing interest in using Natural Language Processing (NLP) techniques for suicide prevention \citep{fernandes2018identifying, bejan2022improving}.
Researchers have developed suicide detection systems to analyze and interpret social media data, including text data. By detecting linguistic markers of distress and other risk factors, these systems can help identify individuals with a risk of suicidality and provide early interventions to prevent such incidents \citep{vioules2018detection}. NLP techniques, therefore, offer a promising avenue for suicide prevention efforts, enabling more proactive and effective interventions to support those in need.

This paper investigates the strengths and limitations of ChatGPT, an advanced language model created by OpenAI \citep{radford2021chatgptplus}, as a tool for suicidal ideation assessment from social media posts.  The ChatGPT API provides access to a powerful natural language processing tool that can generate human-like text, answer questions, and perform a variety of other language-related tasks.
With ChatGPT, developers can build conversational interfaces, Chatbots, and virtual assistants to interact with users and provide informative responses. However, some studies have highlighted the potential risks and ethical concerns associated with the use of ChatGPT and other language models in sensitive domains, such as mental health and suicide prevention \citep{zhuo2023exploring}. Therefore, it is crucial to carefully evaluate the use of ChatGPT  in such settings to better appreciate its potential and limitations.


Our two research questions to assess the reliability of ChatGPT in the suicide prevention task are as follows:
\begin{itemize}[leftmargin=*]
\item\textbf{{RQ1: Can ChatGPT assess the level of suicidality indicated in a written text?}}
\item\textbf{{RQ2: Is ChatGPT's performance comparable with transformer-based models trained on a human-annotated dataset?}}
\end{itemize}

The main contributions of this study are as follows:

\begin{itemize}[leftmargin=*]
\item  Our study examines ChatGPT's performance in evaluating the level of suicidality in Reddit posts using Zero-Shot and Few-Shot Learning techniques. We also compare the results of ChatGPT with two transformer-based models, ALBERT and DistilBERT. Our findings suggest that ChatGPT has potential for suicide risk assessment in Zero-Shot learning, but ALBERT outperforms ChatGPT in this task.

\item  Our study examines how changing temperature parameters affect ChatGPT's ability on suicide risk assessment. We found that the rate of inconclusive responses generated by ChatGPT is closely linked to changes in the temperature parameter, particularly in the Zero-Shot setting.

\item  Based on our findings, we can infer that, at lower temperature values, ChatGPT refrains from making a decision for a greater number of instances; however, it exhibits higher accuracy on the subset of instances for which it makes a decision.
\end{itemize}
These contributions provide a comprehensive evaluation of ChatGPT's performance in this critical application domain and highlight it's potential to be used to assist suicide prevention experts. Our code is available at GitHub\footnote{\url{https://github.com/Hamideh-ghanadian/ChatGPT_for_Suicide_Risk_Assessment_on_Social_Media}}.

\section{Background and Related Work}\label{sec:background}
In this section, we review the related work in suicide ideation detection as well as the generative language technology of the ChatGPT model.   
\subsection{\textbf{Suicidal ideation detection and assessment}} 
There are a vast number of research techniques that investigate suicidal ideation and its cause. For instance, clinical methods examine the resting state of heart rate \citep{sikander2016predicting} and event-related initiators such as depression \citep{jiang2015erp} as suicidal indicators. Traditional methods use questionnaires, electronic health records, and face-to-face interviews to assess the potential risk of suicide \citep{chiang2011socio}.

Several studies indicated the impact of social network reciprocal connectivity on users’ suicidal ideation. \citet{hsiung2007suicide} analyzed the changes in user behavior following a suicide case that occurred within a social media group. \citet{jashinsky2014tracking} 
highlighted the geographic correlation between the suicide mortality rates and the occurrence of risk factors in tweets. \citet{colombo2016analysing} focused on analyzing tweets that contained suicidal ideation, with a particular emphasis on the users' behavior within social network interactions that resulted in a strong and reciprocal connectivity, leading to strengthened bonds between users.

In recent years, NLP researchers have started to analyze users’ posts on social media websites to gain an insight into language usage and linguistic clues of suicidal ideation \citep{Chowdhary2020NaturalLP,babulal2023suicidal}. 
Using NLP techniques, suicide-related keyword dictionaries and lexicons are manually built to enable keyword filtering \citep{Varathan2014SuicideDS}. 
The related analysis contains lexicon-based filtering \citep{sarsam2021lexicon}, topic modeling within suicide-related posts \citep{seah2018data}, transformer-based models, and unsupervised learning \citep{linthicum2019machine}. In line with this field of research, we examine the use of the ChatGPT model for this task, where no labeled data (Zero-Shot setting) or a small labeled dataset (Few-Shot setting) is available. 

\subsection{\textbf{ChatGPT}} ChatGPT is a state-of-the-art artificial intelligence (AI) Chatbot developed by OpenAI \citep{radford2021chatgptplus} that has gained widespread attention for its ability to generate human-like text. 
The original GPT model was trained on a massive corpus of text data, including books, articles, and web pages, using an unsupervised learning approach. The model's performance on a range of language tasks has since been surpassed by newer models, including GPT-2 \citep{radford2019language} and GPT-3 \citep{brown2020language}, which have larger training datasets and more sophisticated architectures.
However, the ChatGPT model has been fine-tuned on large datasets of conversation data, including social media posts, customer support interactions, and chatbot logs \citep{dwivedi2023so}. ChatGPT differs from prior models as it employs Reinforcement Learning from Human Feedback (RLHF). Unlike supervised learning methods that depend on pre-existing training data, RLHF generates a response to a given input, which is evaluated by a human reviewer. The feedback obtained from the evaluator is used to train the model using reinforcement learning, with the objective of maximizing the reward received \citep{lambert2022illustrating}.

Several recent studies have explored the effectiveness of ChatGPT in a variety of settings, including chatbots and virtual assistants. 
One study created a corpus named Human ChatGPT Comparison Corpus (HC3) by collecting a set of question-and-answer datasets covering various domains such as finance, medicine, and psychology \citep{guo2023close}. They conducted a comparative analysis between the responses generated by ChatGPT and those provided by humans to investigate the distinguishing features of ChatGPT's responses. In \citet{jeblick2022chatgpt} ChatGPT was employed to produce a simplified version of a radiology report, which was then evaluated for quality by radiologists. Another study investigated the proficiency of ChatGPT in answering questions related to the United States Medical Licensing Examination (USMLE) Step 1 and Step 2 exams \citep{gilson2022well}. They found that ChatGPT performed similarly to a third-year medical student. 

\citet{bang2023multitask} proposes a framework for evaluating interactive ChatGPT language learning models using publicly available datasets. They evaluated  ChatGPT using 23 datasets covering 8 different NLP tasks, such as summarization, machine translation, sentiment analysis, question answering, etc.
They reported that ChatGPT outperforms large language models with Zero-Shot Learning on most tasks and even outperforms fine-tuned models on some tasks.

In this study, we analyze the performance of ChatGPT in predicting suicidal ideation on social media and identifying possible errors that may occur during the process. 

\section{Dataset}\label{sec:data}
 We utilize the University of Maryland Reddit Suicidality Dataset(UMD) \cite{zirikly2019clpsych,shing2018expert}, which is 
 collected from the Reddit platform. Reddit is an online website and forum for anonymous discussion on a wide variety of topics. It is made up of millions of collective forums or groups called subreddits,
 including the \textit{Depression}\footnote{\href{https://www.reddit.com/r/depression/}{Depression subreddit}} and \textit{SucideWatch}\footnote{\href{https://www.reddit.com/r/SuicideWatch/}{SuicideWatch subreddit}} subreddits. 

 The UMD dataset is a collection of Reddit posts and comments created by individuals who expressed suicidal thoughts or behaviors. 
 The dataset contains over 100,000 posts and comments collected from various subreddits, including those related to mental health and suicide prevention, such as ``\textit{r/SuicideWatch}''. The data was collected over a period of several years and includes the content of the post and comments as well as the location and timing of the posts.

UMD has been repeatedly  used by researchers to develop and test natural language processing algorithms and machine learning models to identify and analyze patterns in online communication related to suicide risk \citep{coppersmith2018predicting}.
\citet{ji2022suicidal} proposed a method for improving text representation through the incorporation of sentiment scores based on lexicon analysis and latent topics. Additionally, they introduce the use of relation networks for the detection of suicidal ideation and mental disorders, leveraging relevant risk indicators.
\citet{ji2021mentalbert} utilized two pretrained masked language models, MentalBERT and MentalRoBERTa, specifically designed to support machine learning in the mental healthcare research field. The authors assess these domain-specific models along with various pretrained language models on multiple mental disorder detection benchmarks. The results show that utilizing language representations pretrained in the mental health domain enhances the performance of mental health detection tasks, highlighting the potential benefits of these models for the mental healthcare research community.

This dataset contains annotations at the user level, utilizing a four-point scale to indicate the severity of the  suicide risk: (a) \textit{No risk}, (b) \textit{Low risk}, (c) \textit{Moderate risk}, and (d) \textit{High risk}. According to \citet{zirikly2019clpsych}, the dataset is divided into three subsets,  each containing annotations for a distinct task. In this study, we utilized  the subset designated for Task A. This task focuses on risk assessment and involves simulating a scenario in which an individual is suspected to require assistance based on online activity, such as posting to a relevant forum or discussion (e.g., r/SuicideWatch). The objective of the task is to evaluate the individual's risk level based on their online activity. This task requires minimal data, with each user typically having posted no more than a few times on SuicideWatch.


 

\vspace{5pt}
\noindent\textbf{Data Preprocessing:}
In this study, we only use a subset of the UMD dataset. This subset of the dataset is designed for a specific task (Task A) and includes posts from 21,518 users and is subdivided into 993 labeled users and 20,525 unlabelled users.
Out of the 993 labeled users, 496 have at least posted once on the SuicideWatch subreddit. The remaining 497 users are control users (i.e., they have not posted on SuicideWatch or any mental health-related subreddits). Since the provided labels are user-level labels, we aggregated all the posts of each user into a single data point, through the concatenation of all the posts made by a particular user. The dataset is divided into 80\% training and 20\% testing subsets. The ChatGPT evaluation was conducted solely on the testing subset, comprising 172 instances with proportional representation for each label. Table~\ref{tab:data} presents the class sizes 
of the data subset used in this project.
 
\begin{table}[ht]
\centering
\resizebox{\columnwidth}{!}{%
\begin{tabular}{ccccc}
\hline
\textbf{}              & \textbf{No Risk} & \textbf{Low Risk} & \textbf{Moderate Risk} & \textbf{High Risk} \\ \hline
\textbf{UMD Dataset} & \textbf{26.73 \%} & \textbf{15.27}  & \textbf{30.69 \%}      & \textbf{27.28 \%}  \\ \hline
\textbf{\# of Users}   & \textbf{196}     & \textbf{112}      & \textbf{225}           & \textbf{200}       \\ \hhline{=#=#=#=#=}
\textbf{Training subset }   & \textbf{27.45 \%}     & \textbf{16.39 \%}      & \textbf{31.90 \%}           & \textbf{24.24 \%}       \\ \hline
\textbf{\# of Users }   & \textbf{154}     & \textbf{92}      & \textbf{179}           & \textbf{136}       \\ \hhline{=#=#=#=#=}
\textbf{Testing subset }   & \textbf{24.41 \%}     & \textbf{11.62 \%}      & \textbf{26.74 \%}           & \textbf{37.20 \%}       \\ \hline
\textbf{\# of Users }   & \textbf{42}     & \textbf{20}      & \textbf{46}           & \textbf{64}       \\ \hline
\end{tabular}%
}
\caption{The description of the subset of the UMD Dataset for TASK A defined in \citet{zirikly2019clpsych}}
\label{tab:data}
\end{table}

\section{Methodology}\label{sec:Method}
This paper evaluates the ability of ChatGPT to predict the level of suicidal ideation on the UMD dataset \citep{zirikly2019clpsych, shing2018expert} and compares it with two fined-tuned classifiers.

\subsection{Fine-Tuned Classifiers}\label{sec:classifiation}
We used pre-trained transformer-based language models to train two text classifiers. Transformers are a class of deep learning models, first introduced by \citet{vaswani2017attention} in 2017.
Researchers build state-of-the-art NLP models using transformer-based architectures because they can be quickly trained on large datasets and studies have shown that they are better at modeling long-term dependencies in natural language text. \citep{wolf2020transformers}. Moreover, the growth of pre-trained transformer-based structures has made it easier to adapt a high-capacity model trained on a large text to downstream tasks \citep{devlin2018bert,howard2018universal}.


We utilize ALBERT \footnote{\href{https://huggingface.co/docs/transformers/model_doc/albert}{AlBERT}} and DistilBERT\footnote{\href{https://huggingface.co/docs/transformers/model_doc/distilbert}{DistilBERT}} language models and fine-tune them with the UMD dataset to build the classifiers. 
For implementation, we employed the Huggingface library \citep{wolf2019huggingface}, an open-source library and data science platform that provides tools to build, train and deploy ML models. 

The ALBERT model was proposed by \citet{lan2019albert}
as a variation of BERT that is optimized in terms of memory consumption and training speed. In other words, ALBERT is a more lightweight version of BERT that maintains its high level of accuracy, making it a powerful tool for various NLP applications.
The DistilBERT model was proposed by \citet{sanh2019distilbert} which has \%40 fewer parameters than BERT and runs \%60 faster while preserving over \%95 of BERT’s performances.

We used the Trainer\footnote{\href{https://huggingface.co/docs/transformers/main_classes/trainer}{Trainer}} class from Huggingface transformers\footnote{\href{https://huggingface.co/docs/transformers/index}{Huggingface Transformers}} for feature-complete training in PyTorch.  
The hyperparameters were selected based on the default values commonly used in similar studies. The final hyperparameters used in our experiments were Learning Rate= $2e^{-5}$, Batch Size = 4, Dropout Rate = 0.1, and Maximum Sequence Length = 512.


\subsection{ChatGPT}\label{sec:chatgpt}
The language model utilized by ChatGPT is \textit{gpt-3.5-turbo}\footnote{\url{https://platform.openai.com/docs/models/gpt-3-5}}, which is one of the most advanced language models developed by OpenAI. Chat models accept a sequence of messages as an input and produce a message generated by the model as an output. Although the chat format is primarily intended for conversations spanning multiple turns, it is also equally useful for single-turn tasks that do not involve any conversations. We used the \textit{OpenAI Python library}\footnote{\url{https://github.com/openai/openai-python}} to access the \textit{ChatCompletion} functionality of the \textit{gpt-3.5-turbo} model through its API. 

\vspace{5pt}
\noindent\textbf{Zero-Shot and Few-Shot Learning:} 
Zero-Shot Learning and Few-Shot Learning are two techniques in Natural Language Processing (NLP) that are used to develop models that can perform tasks without explicit training data \citep{puri2019zero}.
In this project, we evaluate the ability of ChatGPT for Zero-Shot Learning to predict the severity of the suicidality of text for which it has not been explicitly trained. Moreover, we aim to investigate ChatGPT's ability to predict the severity of suicidality, in a Few-Shot Learning setting, where a limited amount of training data is available. In this approach, the model is trained on a small set of examples of the task and then tested on a more extensive set of examples. 

\vspace{5pt}
\noindent\textbf{Input Message:}
The primary input for the system is the 
``message'' parameter, which must be an array consisting of message objects.  
This object includes a ``role'' (either ``system'', ``user'', or ``assistant'') and a ``content'' (the message content).
A conversation can consist of a single message or can extend over multiple pages. We provide a single message to the system which describes the definitions of suicide severity assessment. 

The following example shows the input message for Zero-Shot Learning. In order to 
be consistent with the instructions given to human annotators of the UMD dataset, we obtained the definitions of each risk category 
from \citet{shing2018expert}. These definitions have been adapted from \citet{corbitt2016college} and specify a  
four-way categorization scheme of risk. 

\begin{small}
\begin{lstlisting}[language=Python,breaklines=true] 
res=openai.ChatCompletion.create(
model="gpt-3.5-turbo",
messages=[{"role": "system", "content": 
"""specify the level of the suicidality of the text based on the following criteria: 

0= I do not see evidence that this person is at risk for suicide.

1= There may be some factors here that could suggest risk, but I do not really think this person is at much of a risk of suicide.

2= I see indications that there could be a genuine risk of this person making a suicide attempt.    

3= I believe this person is at high risk of attempting suicide in the near future. Only say number"""},


{"role": "user","content":"""DESIRED INPUT TEXT"""}],
temperature=0.1)        
\end{lstlisting}
\end{small}

For Few-Shot Learning with ChatGPT, we use prompt engineering. The prompt consists of two examples for each category (eight in total) drawn from the training dataset to the input message and followed by the same assessment question. 
For prompt engineering, we drew inspiration from a short course on ChatGPT Prompt Engineering\footnote{\href{https://www.deeplearning.ai/short-courses/chatgpt-prompt-engineering-for-developers/}{ChatGPT Prompt Engineering for Developers}} offered by \textit{DeepLearning.AI}. We initiated the prompt construction process with a simple initial prompt and iteratively refined it through multiple rounds of trial and error. This iterative approach allowed us to gradually evolve the prompt, making necessary adjustments based on the observed outcomes and performance of the model. 
The complete implementation including Zero-Shot Learning, Few-Shot Learning and the fine-tuned classifiers is available on GitHub\footnote{\url{https://github.com/Hamideh-ghanadian/ChatGPT_for_Suicide_Risk_Assessment_on_Social_Media}}.

\vspace{5pt}
\noindent\textbf{Temperature Parameter:}
The Temperature value in  ChatGPT is a parameter that controls the randomness and creativity of the model's responses. To produce a response to a given input message, the model generates a probability distribution over all possible next words or tokens in the response. The temperature parameter affects the probability distribution over the possible tokens at each step of the generation process.

A high temperature value (close to 1) will result in more diverse and unpredictable responses, as the model samples from less likely tokens in the distribution. This can result in more creative and surprising responses but may also increase the likelihood of generating nonsensical or irrelevant text.
On the other hand, a low temperature value (e.g. 0.1) will result in more conservative and predictable responses, as the model chooses the most likely tokens in the distribution. This can result in more coherent and on-topic responses but may be more repetitive or less attractive.
The temperature parameter in the ChatGPT allows users to control the balance between creativity and coherence in the model's responses based on their specific needs and preferences. 

\vspace{5pt}
\noindent\textbf{Inconclusiveness Rate:} We define an additional metric, the \textit{Inconclusiveness rate} for further evaluation of ChatGPT in this task. This parameter refers to the 
proportion of test cases that do not yield a definitive or conclusive result. In other words, it is the rate at which the evidence or information is  inconclusive to support a clear decision. 
To calculate the inconclusiveness rate, after ChatGPT assessed the suicidality risk level of instances, we count all the cases where the ChatGPT reports inconclusive results. 
An example of an inconclusive response generated by ChatGPT is, \textit{``As an AI, I cannot provide an assessment of the suicidal risk level for this instance''}. 
Then we divide the number of inconclusive instances by the total number of instances in the test dataset and report this metric as a percentage.

\section{Results}\label{sec:Results}
In this section, we present the results of our study in accordance with the research questions presented in section \ref{sec:intro} based on the test set 
described in section \ref{sec:data}. For evaluation,  we report four widely-used metrics in this task, accuracy,  precision, recall, and F-score
to provide a comprehensive and informative evaluation of the performance of the classification models \citep{sokolova2009systematic}. For ChatGPT, we also report the  \textit{Inconclusiveness rate} described in Section \ref{sec:Method}. 

\subsection{RQ1: Can ChatGPT assess the level of suicidality indicated in a written text?}

\vspace{5pt}
\noindent\textbf{Zero-Shot Learning:} 
In this section, we present the results of ChatGPT for suicidal ideation prediction with the Zero-Shot Learning approach. The goal of our project is to evaluate the performance of ChatGPT in assessing the level of suicidality of a written text. Furthermore, we use five different temperature values to evaluate the impact of temperature on generated response, and report the inconclusiveness rate of ChatGPT at each temperature. The rest of the metrics are used to evaluate the performance of ChatGPT for the instances in that ChatGPT was able to generate a conclusive answer. Table \ref{tab: Zero shot} presents the performance of the ChatGPT in five different temperature values. 

\begin{table}[ht]
\resizebox{\columnwidth}{!}{%
\begin{tabular}{cccccc}
\hline
\textbf{Temperature} & \textbf{Accuracy} & \textbf{Precision} & \textbf{Recall} & \textbf{F1-Score} & \textbf{Inconclusiveness Rate} \\ \hline
\rowcolor{yellow!50}
\textbf{0.1} & \textbf{0.88} & \textbf{0.57} & \textbf{1}    & \textbf{0.73} & \textbf{2.91 \%}  \\
\textbf{0.3} & \textbf{0.67} & \textbf{0.33} & \textbf{1}    & \textbf{0.50} & \textbf{2.32 \%}  \\
\textbf{0.5} & \textbf{0.67} & \textbf{0.22} & \textbf{0.67} & \textbf{0.33} & \textbf{1.71 \%} \\
\textbf{0.7} & \textbf{0.64} & \textbf{0.27} & \textbf{1}    & \textbf{0.43} & \textbf{1.16 \%}  \\
\textbf{1}   & \textbf{0.54} & \textbf{0.21} & \textbf{1}    & \textbf{0.35} & \textbf{0 \%}  \\ \hline
\end{tabular}%
}
\caption{Performance and inconclusiveness rate of ChatGPT for Zero-Shot Learning in five different temperature values. The row with the highest F1-score is highlighted.}
\label{tab: Zero shot}
\end{table}

\begin{figure*}[ht]
    \centering
    \begin{subfigure}[b]{0.32\linewidth}
        \includegraphics[width=\linewidth]{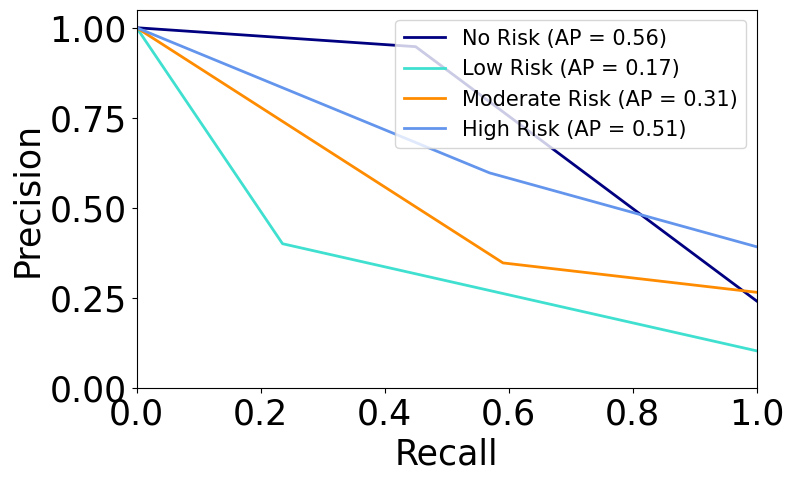}
        \captionsetup{font=tiny}
        \caption{Temp=0.1}
    \end{subfigure}
    \begin{subfigure}[b]{0.32\linewidth}
        \includegraphics[width=\linewidth]{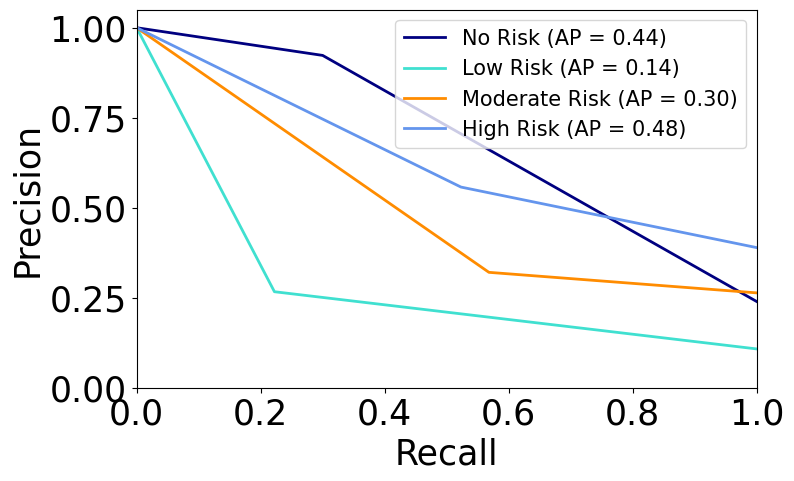}
        \captionsetup{font=tiny}
        \caption{Temp=0.3}
    \end{subfigure}
    \begin{subfigure}[b]{0.32\linewidth}
        \includegraphics[width=\linewidth]{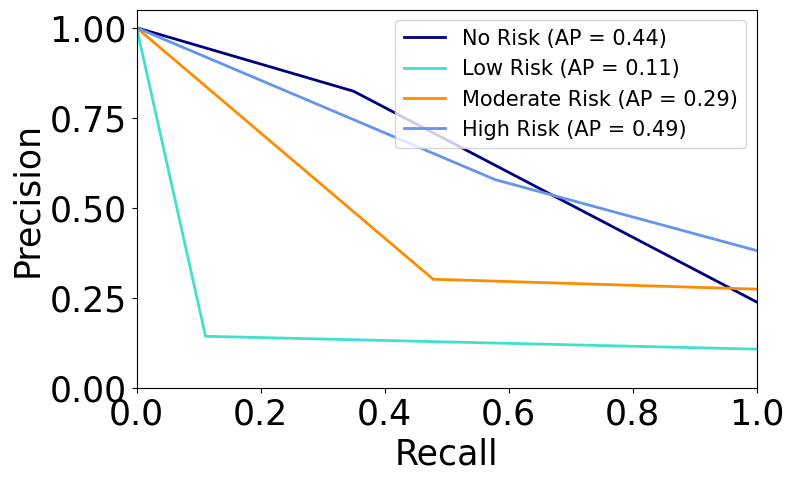}
        \captionsetup{font=tiny}
        \caption{Temp=0.5}
    \end{subfigure}
    \begin{subfigure}[b]{0.32\linewidth}
        \includegraphics[width=\linewidth]{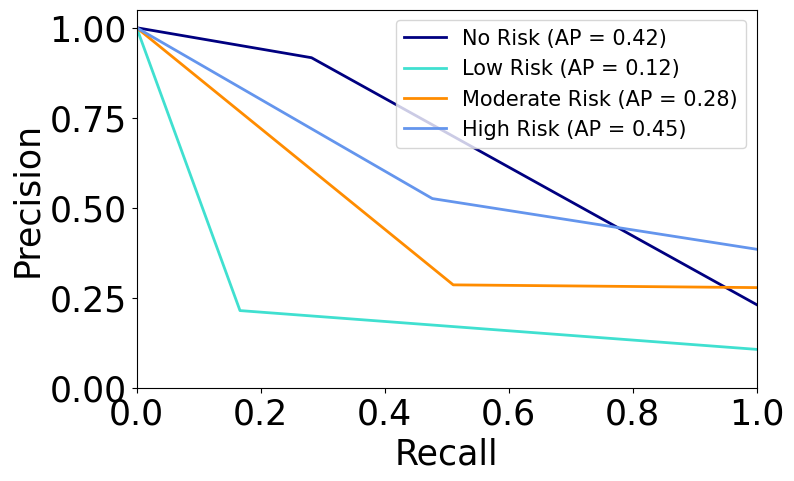}
        \captionsetup{font=tiny}
        \caption{Temp=0.7}
    \end{subfigure}
    \begin{subfigure}[b]{0.32\linewidth}
        \includegraphics[width=\linewidth]{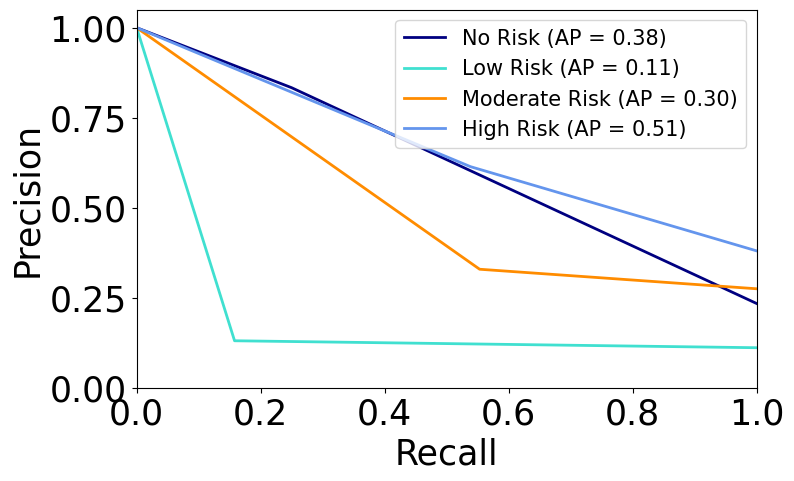}
        \captionsetup{font=tiny}
        \caption{Temp=1}
    \end{subfigure}
    \caption{Precision-Recall graph of the ChatGPT at different temperature values
    in Zero-Shot setting}
    \label{zeroshot PR}
\end{figure*}


As presented in Table \ref{tab: Zero shot}, a higher temperature will result in a more decisive 
output but with a greater risk of generating errors. 
Conversely, a lower temperature will result in more indecisiveness, 
but with a lower risk of errors, i.e., the highest F1-score is achieved with a temperature of 0.1.  
We observed that ChatGPT's inconclusiveness rate (inability to assess the level of suicidality of instances) is 2.91\% for a temperature of 0.1, which is the highest rate for all temperature values. As shown in Table \ref{tab: Zero shot}, as the temperature value increases, the inconclusiveness rate and F1-score decrease.

\begin{figure}[htbp]
\centering
    {\includegraphics[width=0.8\linewidth]{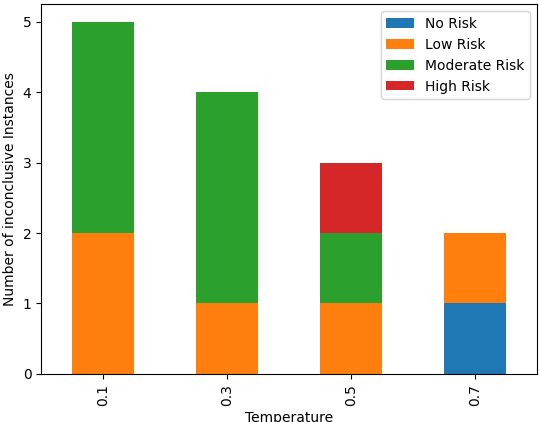}}
    \caption{ 
    Number of instances for which ChatGPT refrains from making a decision, at different temperature values and for classes \textit{No Risk}, \textit{Low Risk}, \textit{Moderate Risk}, and \textit{High Risk} }
    \label{inconclusiveness rate bar}
\end{figure}

For further evaluation, we present the Precision-Recall (PR) graph of the model at each temperature. 
The PR graph displays the trade-off between precision and recall for different thresholds used to classify instances.
Figure \ref{zeroshot PR} shows the PR curve of ChatGPT for each class. Moreover, it shows the impact of increasing temperature values on predicting the suicidality of the text in each class. As the temperature increases, the area under the PR graph declines. In other words, the graph shows lower values for both precision and recall measures. Moreover, Figure \ref{zeroshot PR} shows that the Average Precision (AP) of ChatGPT in predicting the \textit{No Risk} and \textit{High Risk} classes is higher,  compared to the two middle classes, \textit{Low Risk} and \textit{Moderate Risk}.

The bar chart depicted in Figure \ref{inconclusiveness rate bar} illustrates which classes are more challenging for ChatGPT for suicidality assessment. 
Figure \ref{inconclusiveness rate bar} shows that at the temperature of 0.1, 3 out of 5 inconclusive instances belong to \textit{Moderate risk} and 2 out of 5 instances belong to \textit{Low Risk} categories.

\vspace{5pt}
\noindent\textbf{Few-Shot Learning:} 
We use prompt engineering to implement Few-Shot Learning with ChatGPT.
The prompt consists of a few  examples from the training dataset, and the model is trained to assess the suicidality level of the text based on the given criteria. Similar to Zero-Shot Learning, we begin by providing the definitions of each risk category in the prompt, followed by eight training examples and their corresponding labels, with each example and label being separated by a comma and placed in individual paragraphs. The prompt concludes with a request for ChatGPT to provide an assessment based on the given criteria.
Table~\ref{tab: Few Shot} presents the results of ChatGPT in Few-Shot settings at different temperature values.

\begin{table}[ht]
\resizebox{\columnwidth}{!}{%
\begin{tabular}{cccccc}
\hline
\textbf{Temperature} & \textbf{Accuracy} & \textbf{Precision} & \textbf{Recall} & \textbf{F1-Score} & \textbf{Inconclusiveness Rate} \\ \hline
\rowcolor{yellow!50}
\textbf{0.1} & \textbf{0.81} & \textbf{0.67} & \textbf{0.77}    & \textbf{0.71} & \textbf{0.58 \%}  \\
\textbf{0.3} & \textbf{0.81} & \textbf{0.67} & \textbf{0.77}    & \textbf{0.71} & \textbf{0.58 \%}  \\
\textbf{0.5} & \textbf{0.76} & \textbf{0.57} & \textbf{0.67} & \textbf{0.65} & \textbf{0.58 \%} \\
\textbf{0.7} & \textbf{0.75} & \textbf{0.56} & \textbf{0.77}    & \textbf{0.62} & \textbf{0 \%}  \\
\textbf{1}   & \textbf{0.75} & \textbf{0.56} & \textbf{0.77}    & \textbf{0.62} & \textbf{0 \%}  \\ \hline
\end{tabular}%
}
\caption{Performance of ChatGPT for Few-Shot Learning in five different temperature values. The row with the highest F1-score is highlighted.}
\label{tab: Few Shot}
\end{table}

ChatGPT achieves the highest F1-score at the temperature of 0.1. Furthermore, we observed that the inconclusiveness rate of ChatGPT in Few-Shot Learning was significantly lower compared to Zero-Shot Learning. Additionally, the inconclusiveness rate remained almost constant at different temperature values, indicating that ChatGPT is more confident in generating responses 
when it is provided with a few examples.\medskip
 
\begin{figure}[htbp]
    \centering
    \begin{subfigure}[b]{0.49\linewidth}
        \includegraphics[width=\linewidth]{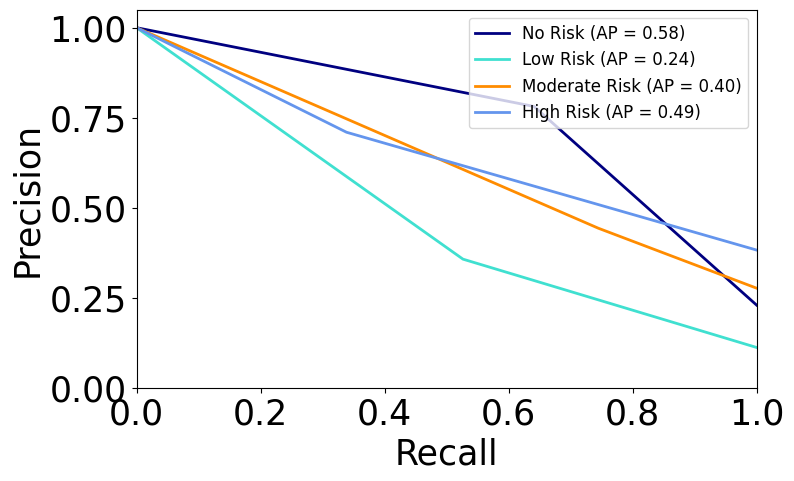}
        \captionsetup{font=tiny}
        \caption{Temperature=0.1}
    \end{subfigure}
    \hfill
    \begin{subfigure}[b]{0.49\linewidth}
        \includegraphics[width=\linewidth]{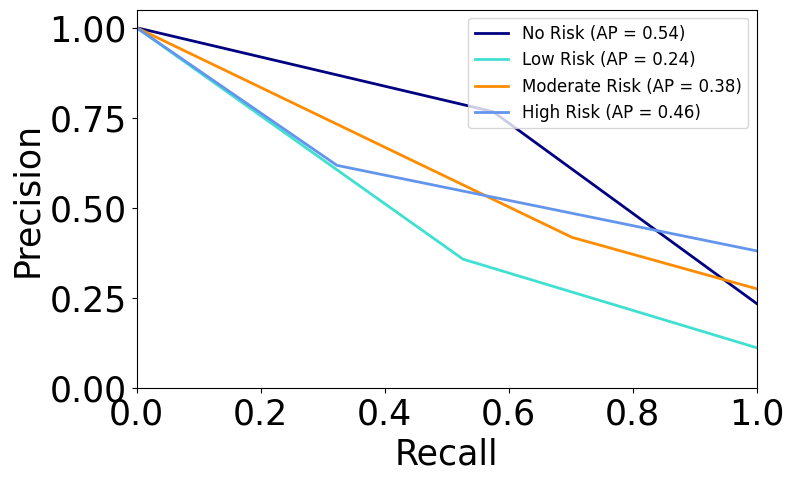}
        \captionsetup{font=tiny}
        \caption{Temperature=1}
    \end{subfigure}
      \caption{Precision-Recall graph of the ChatGPT at two extreme temperature values (0.1 and 1) 
      in a Few-Shot settings, for classes \textit{0=No Risk}, \textit{1=Low Risk}, \textit{2=Moderate Risk}, \textit{3= High Risk}.}
     \label{Fewshot PR}
\end{figure}

Figure \ref{Fewshot PR} presents the 
PR graph of ChatGPT for two extreme temperature values.As presented in Table \ref{tab: Few Shot}, the precision and recall values for temperature values 0.1 and 1 are not significantly different as it is reflected in the PR curve as well. However, the PR curve of two classes, \textit{Moderate Risk} and \textit{High Risk}, slightly improves by decreasing the temperature.

\subsection{RQ2: Is ChatGPT's performance comparable with transformer-based models trained on a human-annotated dataset?} To train a classification model on the UMD datasets, we employed two pretrained transformer-based models, DistilBERT and ALBERT. The 
performances of these models on the aforementioned dataset are presented in Table~\ref{tab: performance comparison} and are compared with the results obtained by the ChatGPT model.

\begin{table}[ht]
\resizebox{\columnwidth}{!}{%
\centering
\begin{tabular}{ccccc}
\hline
\textbf{Model}      & \textbf{Accuracy} & \textbf{Precision} & \textbf{Recall} & \textbf{F1-Score} \\ \hline
\textbf{AlBERT}     & \textbf{0.865}    & \textbf{0.861}     & \textbf{0.865}  & \cellcolor{yellow!50}\textbf{0.869}    \\
\textbf{DistilBERT} & \textbf{0.77}     & \textbf{0.804}     & \textbf{0.771}  & \textbf{0.745}    \\ \hline
\textbf{Zero-Shot ChatGPT (temp=0.1)} & \textbf{0.88} & \textbf{0.57} & \textbf{1} & \textbf{0.73} \\
\textbf{Few-Shot ChatGPT (temp=0.1)} & \textbf{0.81} & \textbf{0.67} & \textbf{0.77} & \textbf{0.71} \\ \hline
\end{tabular}%
}
\caption{Comparison of the two transformer-based models with ChatGPT. Fine-tuned ALBERT is highlighted for achieving the highest F-score.}
\label{tab: performance comparison}
\end{table}

As presented in Table \ref{tab: performance comparison}, while ChatGPT's performance is comparable to a fine-tuned DistillBERT, it falls considerably short (by 13\% for F1-score) compared to a fine-tuned ALBERT model.

\section{Discussions and Conclusion}\label{sec:Discussion}
This study focuses on the evaluation of the accuracy and quality of response generated by ChatGPT for the assessment of suicidal ideation levels. The performance of ChatGPT was assessed in  Zero-Shot and Few-Shot Learning scenarios. Zero-Shot Learning can be particularly useful when obtaining labeled data is difficult or expensive.
In Zero-Shot Learning, ChatGPT achieved an F1-score of 0.73, 
on our test set (temperature=0.1). These findings demonstrate the potential of ChatGPT as a tool for data annotation, particularly when utilizing a simple prompt design. However, it is important to note that in sensitive tasks such as suicidal ideation detection or assessment, caution must be exercised to ensure accuracy and ethical considerations should be prioritized. 

We conducted a Few-Shot Learning experiment to assess the performance of ChatGPT when a few labeled examples of the training data are appended to the prompt. We achieved an F1-Score of 0.71 in Few-Shot Learning (temperature=0.1).
In Zero-Shot Learning, the model is able to leverage its existing knowledge to make predictions for new tasks. This approach can be particularly effective when the model needs to generalize to a wide range of possible new tasks. On the other hand, Few-Shot Learning requires the model to learn from a limited amount of training data for each new task. This approach can be more challenging, as the model has to generalize from a small set of examples and may struggle to identify patterns or relationships that are important for the new task.

In this study, we carried out an experiment to examine the impact of the temperature hyperparameter on the performance of ChatGPT. In Zero-Shot Learning, our findings indicate that there is a negative correlation between the F1-Score and the temperature hyperparameter. In other words, as the temperature increases, the model's performance tends to decrease. These results suggest that careful optimization of hyperparameters, such as temperature, is crucial for achieving optimal performance of ChatGPT. In Few-Shot Learning, there is still a negative correlation between the F1-Score and the temperature hyperparameter. However, the change in the F1-Score value is subtle, indicating that the impact of temperature tuning on model performance may not be significant.

Another discovery highlighted in this paper pertains to the examination of the inconclusiveness rate of ChatGPT. There is a trade-off between the inconclusiveness rate and the F1-score in order to optimize the performance of the ChatGPT. In sensitive tasks such as suicide assessment risk, it is crucial to have a highly accurate model that can provide reliable predictions. In some cases, it may be preferable for the model to provide an \textit{``I do not know'' } response rather than providing unreliable predictions about suicidality. Careless responses from a suicidal assessment model can have serious consequences, including false positives or false negatives, which can harm individuals at risk. Table \ref{tab: Zero shot} shows that the inconclusiveness rate in temperature 0.1 of Zero-Shot Learning is 2.91\%,  and the F1-Score is 0.73. By increasing the temperature, we have fewer inconclusive instances and yet a lower F1-score over the rest of the responses. Table \ref{tab: Few Shot} for Few-Shot Learning  shows that the \textit{inconclusiveness rate} becomes almost constant and smaller among different temperature values because the model has learned to generalize based on the limited number of examples provided during training and the model is not able to generate as much variation in response because it may over-fit to the training examples. As a result, the model may be less prone to generating random or unexpected responses.

Figure \ref{inconclusiveness rate bar} indicates that the inconclusive instances mostly belong to two middle  classes \textit{Low Risk} and \textit{Moderate Risk}. 
These two classes are highly subjective due to the vague boundaries of definitions. For example, the Zero-Shot Learning model was not able to provide an assessment for the following instance: ``\textit{I have ups and downs, I've had them for a long time and I don't know why, since December I've been going to therapy, I've been getting meds too and at first they helped suppress the storm of thoughts that won't let me sleep eat and think, I keep finding myself trying to sleep''}. On the other hand, The Few-Shot Learning predicts \textit{High Risk} suicidality level for this instance, and the human experts annotated this instance as \textit{Low Risk}. This example clarifies that generating an "I do not know" answer here can be preferable to a wrong assessment.

To evaluate how well ChatGPT performs compared to other transformer-based models, we conducted an experiment where we 
fine-tuned two other models, ALBERT and DistilBERT, with the train set of the UMD dataset. The results of this experiment, shown in Table \ref{tab: performance comparison}, suggest that the ALBERT model reaches promising results with an F1-score of 0.869, outperforming both the DistilBERT and ChatGPT models, with F1-scores of 0.745 and 0.73, respectively. While the ALBERT model achieved the highest score among the three models, it should be noted that it is trained on the UMD dataset for the suicidal assessment task specifically. On the other hand, ChatGPT is trained on a large corpus of text data using a self-supervised learning approach for multiple tasks. 

Data collection and annotation are essential but expensive processes in supervised machine learning. Obtaining high-quality labels can be specifically costly and time-consuming in sensitive tasks such as suicide detection. Based on our results, one possible approach to reduce the cost and increase the quality of data annotation is to use ChatGPT in an expert-in-the-loop setting. ChatGPT can assist a human annotator in providing faster and more accurate feedback for a given task. For example, in the case of suicide detection, a human annotator can use ChatGPT to generate responses to various prompts related to suicidal behavior. The annotator can then review the model's output and provide corrections or feedback to refine the output.

\section{Future works and Potentials}
To ensure the effectiveness and fairness of suicide detection using ChatGPT, it is vital to address biases and generalization issues. Conversational models such as ChatGPT are trained on vast amounts of text data, which may contain biases and reflect societal prejudices. Future research should focus on developing bias mitigation techniques to prevent the model from perpetuating harmful stereotypes or stigmatizing individuals. Additionally, efforts should be made to enhance the generalization capabilities of the model by training it on diverse datasets encompassing various demographics, cultures, and languages. This will enable the model to better understand and identify suicidal ideation across different populations. 

Another area for future research is the evaluation 
of other Chatbots, especially the open-source conversational models. For instance, The Open Assistant project \footnote{\url{https://open-assistant.io/}}, developed by LAION-AI, aims to offer a highly capable chat-based large language model to a wide audience. Through extensive training on diverse text and code datasets, it has acquired versatile capabilities such as answering queries, generating text, translating languages, and even producing creative content. Moreover, Vicuna \footnote{\url{https://lmsys.org/blog/2023-03-30-vicuna/}} is an advanced chatbot developed by fine-tuning the Large Language Model Meta AI (LLaMA) using user conversations sourced from ShareGPT. Vicuna is an auto-regressive language model designed to provide natural and immersive conversational experiences which generates highly detailed and well-structured responses, comparable in quality to ChatGPT. By utilizing different models, researchers can contribute to advancing the field of conversational models and unlock their full potential in various applications and domains. 

\section{Limitations}
Our study has several limitations that should be acknowledged. First, the study was conducted on a relatively small test dataset. Future work is needed to assess whether our results are generalizable to larger datasets. Second, while we employed a rigorous methodology for evaluating ChatGPT's performance, we have not measured other safety criteria, such as biases or privacy issues in using this model. 
Third, our study focused only on the initial step of suicide risk assessment and did not explore the use of ChatGPT in ongoing monitoring or intervention.
Fourth, we are unsure if the UMD dataset has been used in the training of ChatGPT in any capacity  since the specifics of the training data of ChatGPT are not disclosed to the public. Future work should focus on creating new datasets to assess the performance of ChatGPT on fully unknown test sets.

It is important to note that despite these limitations, our work represents an important first step in understanding the potential for ChatGPT in suicide risk assessment. Future research should aim to address these limitations and explore the feasibility, safety and effectiveness of ChatGPT in broader clinical settings.

\section{Ethical Considerations}\label{sec:Ethics}
For this research, we obtained ethics approval from the research ethics board at the University of Ottawa. Moreover, The UMD dataset was used with authorization from its creators, and we adhered to the terms of use and ethical standards  \footnote{\href{http://users.umiacs.umd.edu/~resnik/umd_reddit_suicidality_dataset.html}{The University of Maryland Reddit Suicidality Dataset}} provided by them.

The use of ChatGPT for suicide risk assessment raises several ethical considerations.
Firstly, there is the issue of 
safety and reliability. While ChatGPT has shown promise in natural language processing tasks, it is not infallible and can make mistakes or generate false responses. Due to the sensitivity of the suicide detection task, these errors might lead to severe harm to individuals at risk. Therefore, it is important to 1) thoroughly test and validate the accuracy of the model before using it for suicide risk assessment and 2) deploy it in an expert-in-the-loop setting.

Secondly, there is the issue of privacy and confidentiality. Suicide risk assessment might involve sensitive personal information, and there is a risk that the information processed by the ChatGPT could be mishandled or disclosed to unauthorized parties. It is important to ensure that proper security measures are in place to protect the privacy of individuals who interact with the ChatGPT. Automatic de-identification of data before feeding it to ChatGPT could be a potential solution, but it will bring in its own limitations. In any case, obtaining user consent is crucial before engaging in the automatic processing of data by ChatGPT. It is essential to respect individuals' privacy and ensure that they have given their explicit permission before their data is collected, processed, or shared. 

Thirdly, there is the issue of potential psychological harm. Suicide risk assessment can be a sensitive and emotional topic. There is a risk that individuals whose data is assessed by ChatGPT could experience distress or other negative emotions due to the assessment. It is important to have appropriate support mechanisms in place, such as access to mental health professionals or crisis hotlines, to assist individuals who may be in distress.

\nocite{*} 
\bibliography{Chat_GPT}
\bibliographystyle{acl_natbib}

\end{document}